\documentclass[final,1p,times]{elsarticle}

\usepackage{amssymb}
\usepackage{float}
\usepackage{graphicx}
\usepackage{multirow}
\usepackage[numbers,sort&compress]{natbib}

\usepackage{amsmath}

\usepackage{lineno}
\usepackage{multirow}        
\usepackage{graphicx}

\usepackage{hyperref}
\hypersetup{hypertex=true,
	colorlinks=true,
	linkcolor=blue,
	anchorcolor=blue,
	citecolor=blue}

\journal{Neural Networks}

\begin{document}

\begin{frontmatter}




\cortext[cor1]{Corresponding author}
\cortext[cor2]{Corresponding author}

\title{Enhancing Physics-Informed Neural Networks with a Hybrid Parallel Kolmogorov-Arnold and MLP Architecture}

\author[1]{Zuyu Xu}
\author[1]{Bin Lv}
\author[1]{Jiale Wang}
\author[1]{Yuhang Pan}
\author[1]{Jun Wang \corref{cor1}}
\ead{21225@ahu.edu.cn}
\author[1]{Yunlai Zhu}
\author[1]{Zuheng Wu \corref{cor2}}
\ead{wuzuheng@ahu.edu.cn}
\author[1]{Yue hua Dai}

\affiliation[1]{organization={School of Integrated Circuits},
            addressline={Anhui University}, 
            city={Hefei},
            postcode={230601}, 
            state={Anhui},
            country={China}}

\begin{abstract}
Neural networks have emerged as powerful tools for modeling complex physical systems, yet balancing high accuracy with computational efficiency remains a critical challenge in their convergence behavior. In this work, we propose the Hybrid Parallel Kolmogorov-Arnold Network (KAN) and Multi-Layer Perceptron (MLP) Physics-Informed Neural Network (HPKM-PINN), a novel architecture that synergistically integrates parallelized KAN and MLP branches within a unified PINN framework. The HPKM-PINN introduces a scaling factor $\xi$, to optimally balance the complementary strengths of KAN's interpretable function approximation and MLP's nonlinear feature learning, thereby enhancing predictive performance through a weighted fusion of their outputs. Through systematic numerical evaluations, we elucidate the impact of the scaling factor $\xi$ on the model's performance in both function approximation and partial differential equation (PDE) solving tasks. Benchmark experiments across canonical PDEs, such as the Poisson and Advection equations, demonstrate that HPKM-PINN achieves a marked decrease in loss values (reducing relative error by two orders of magnitude) compared to standalone KAN or MLP models. Furthermore, the framework exhibits numerical stability and robustness when applied to various physical systems. These findings highlight the HPKM-PINN's ability to leverage KAN's interpretability and MLP's expressivity, positioning it as a versatile and scalable tool for solving complex PDE-driven problems in computational science and engineering.
\end{abstract}

   

\begin{keyword}
Kolmogorov-Arnold networks \sep Physics-informed neural networks \sep Partial differential equations \sep Networks parallel architecture
\end{keyword}

\end{frontmatter}



\section{Introduction}\label{sec1}
Neural networks have revolutionized numerous fields, enabling significant advancements in applications from computer vision, and natural language processing to complex scientific computations \cite{b1,b2,b3}. Within this landscape, PINNs have become particularly notable, where physical laws, typically represented by PDEs, are embedded directly into the neural network architecture. PINNs leverage neural networks as universal approximators, incorporating PDE residuals into the loss function to approximate solutions through optimization \cite{b4,b5,b6,b7}. Though introduced by Dissanayake et al. in 1994 \cite{b5}, PINNs only recently gained momentum due to improvements in automatic differentiation and computational resources \cite{b8,b9}. Raissi et al. showed how PINNs simplify PDE solving by directly representing differential operators within neural networks, eliminating the need for mesh generation, and offering a computationally efficient alternative for a variety of applications \cite{b10}.

Despite their promise, PINNs face significant challenges that hinder their practical deployment, particularly in terms of computational efficiency and training stability \cite{b11,b12}.Training PINNs can be computationally intensive and unstable, with these demands limiting their scalability—especially when dealing with problems that involve complex multi-scale features or require fast convergence \cite{b13,b14}. To address these limitations, researchers have proposed various methods to accelerate PINN convergence while preserving accuracy \cite{b15,b16,b17,b18,b19,b20,b21,b22,b23}. For instance, Jagtap et al. introduced the Conservative Physics-Informed Neural Network (CPINN) \cite{b22}, a domain decomposition approach that divides the spatial domain into subdomains, each assigned to an independent neural network. This strategy enables parallel computation across nodes, significantly enhancing efficiency in computationally intensive applications. Jagtap et al. later developed the Extended Physics-Informed Neural Network (XPINN) \cite{b23}, which generalizes CPINN’s domain decomposition to accommodate a wider class of PDEs and time discretization. Both CPINN and XPINN leverage parallelization to improve algorithmic efficiency, yet they primarily optimize computational resource allocation without addressing potential enhancements in the expressive capacity of PINNs \cite{b21}. 

Recent efforts to enhance the versatility of PINNs have led to innovative designs, such as the Hybrid Quantum Physics-Informed Neural Network (HQPINN) proposed by Sedykh et al \cite{b24}. HQPINN integrates quantum and classical network components, enabling specific sections of the input vector to be processed by either quantum or classical network segments \cite{b25,b26}. This hybrid approach improves the expressive power of PINNs, allowing them to learn diverse patterns and enhancing robustness, particularly in fluid dynamics applications \cite{b27,b28}. However, HQPINN’s enhancements are mainly tailored to specialized problem types, highlighting the need for a more general solution capable of addressing a wider range of PDEs.   

Among the most commonly used neural network architectures for solving PDEs is the fully connected MLP, celebrated for its universal approximation capabilities \cite{b29,b30}. Despite its strengths, MLPs face challenges such as parameter redundancy, limited interpretability, and spectral bias, which can hinder their effectiveness in PINNs \cite{b31,b32}. In contrast, KAN based on the Kolmogorov-Arnold representation theorem, offers a promising alternative. This theorem asserts that any multivariable continuous function can be expressed through compositions of univariate functions \cite{b33}. Hecht-Nielsen’s introduction of the Kolmogorov Network \cite{b34,b35}, later extended by Liu et al \cite{b36}. has proven effective in approximating complex functions. Enhanced with deep architectures and alternative activation functions such as Chebyshev polynomials \cite{b37}, radial basis functions \cite{b38}, and wavelets \cite{b39}, KAN excels at capturing nonlinear dynamics \cite{b36}. However, recent studies indicate that KAN struggles with extracting low-frequency components, which are essential for applications requiring the representation of smooth variations \cite{b40,b41,b50}.

The complementary strengths of MLP and KAN networks provide a compelling foundation for a hybrid approach. MLPs excel in learning across a broad frequency spectrum, while KAN is adept at capturing nonlinear features but lacks proficiency in low-frequency feature extraction \cite{b36,b42,b43}. To address these limitations, we propose a novel hybrid architecture: Physics-Informed Neural Networks with Hybrid Parallel KAN and MLP (HPKM). The HPKM architecture employs a tunable scaling factor $\xi$ , to dynamically adjust the output balance between MLP and KAN, thereby optimizing performance. Our experimental results underscore the advantages of the HPKM architecture. When applied to benchmark PDE problems, such as the Poisson and Advection equations, HPKM consistently achieves smaller loss values compared to standalone MLP or KAN models. This superior performance highlights its ability to capture both high-frequency details and smooth, low-frequency features, addressing the limitations of conventional PINNs. Moreover, HPKM exhibits enhanced convergence rates, improved stability, and robust generalization across diverse physical scenarios, making it a versatile, and efficient solution for challenging computational problems.

The outline of the paper is as follows. Section 2 introduces the fundamental principles of PINNs, provides an overview of KAN networks, and details the construction of the proposed HPKM model. Section 3 presents numerical experiments, focusing on two aspects: (1) Multiscale function fitting, emphasizing challenges involving mixed high- and low-frequency components, and (2) Solving PDEs, including the Poisson equation, Advection equation, Convection-Diffusion equation, and two-dimensional Helmholtz equation. We explore the impact of the tunable parameter $\xi$ on network performance, and compare the HPKM against standalone MLP and KAN models. Finally, Section 4 summarizes the current advantages and limitations of HPKM, providing insights for future work.

\section{Methord}\label{sec2}
\subsection{PINN}\label{subsec21}

PINNs are advanced models designed for solving PDEs by leveraging the capabilities of modern deep learning frameworks. These frameworks allow for the systematic construction of problem-specific loss functions using automatic differentiation techniques \cite{b9}. PINNs approximate the solution of PDEs by iteratively updating the weights and biases of the neural network via gradient descent, thereby minimizing a composite loss function \cite{b10}.

We consider a general PDE system with the following initial and boundary conditions:
\begin{equation}
	\left\{
	\begin{aligned}
		&u_t(x, t) + N[u_t(x, t)] = 0 & x \in \Omega, \, t \in [0,T] \\
		&u(x, 0) = g(x)& x \in \Omega \\
		&B[u] = 0 & t \in [0,T], \, x \in \partial \Omega
	\end{aligned}
	\right.
\end{equation}

where $x$ and $t$ represent the spatial and temporal coordinates, respectively, and $u(x,t)$ denotes the solution of the PDE. The function $g(x)$ represents the initial state of the solution $u$. The subscript $t$ indicates the partial derivative of the function $u(x, t)$ with respect to time, $N[\cdot]$ denotes a linear or nonlinear differential operator, and $B[\cdot]$ represents the boundary condition operator. The domain $\boldsymbol{\Omega}$ is a subset of $\mathbb{R}^{D}$.

The primary objective of PINNs is to approximate the theoretical solution $u(x,t)$ of the PDE system by formulating loss functions based on the governing equations and initial/boundary conditions (I/BC). The residual for the PDE can be expressed as: $R_\theta(x,t)=u_t(x_R^i,t_R^i)+N[u_t(x_R^i,t_R^i)]$, where $\mathcal{R}_\theta(x,t)$ quantifies the deviation from the true solution, and a desirable solution minimizes $\mathcal{R}_\theta(x,t)$ to approach zero. quantifying the deviation from the actual PDE. Thus, it is desirable for its value to approach zero.

In this study, the PINN is trained in an supervised manner, utilizing only losses derived from initial and boundary conditions. These losses are defined as follows:
\begin{equation}
	L_{IC} = \frac{1}{N_{IC}} \sum_{i=1}^{N_{IC}} \left| u_\theta(x^{i}_{IC}, 0) - g(x^{i}_{IC}) \right|^2
\end{equation}
\begin{equation}
	L_{BC} = \frac{1}{N_{BC}} \sum_{i=1}^{N_{BC}} \left| B[u_\theta](x^{i}_{BC}, t^{i}_{BC}) \right|^2
\end{equation}
\begin{equation}
	L_{R} = \frac{1}{N_{R}} \sum_{i=1}^{N_{R}} \left| u_t(x^{i}_{R}, t^{i}_{R}) + N[u_\theta(x^{i}_{R}, t^{i}_{R})] \right|^2
\end{equation}

where $\theta$ represents the network parameters to be optimized, and $u_{\theta(x,t)}$ is the predicted solution by the PINN. In the equation, 
$\left\{x_{IC}^i, 0\right\}_{i=1}^{N_{IC}}$, 
$\left\{x_{BC}^i, t_{BC}^i\right\}_{i=1}^{N_{BC}}$ and 
$\left\{x_R^i, t_R^i\right\}_{i=1}^{N_R}$. denote the initial condition points, boundary condition points and configuration points, respectively. Various sampling methods, such as Latin hypercube sampling \cite{b44} and Sobol sequences \cite{b45}, can be utilized to enhance the convergence of PINN during training. $N_{IC}$,$N_{BC}$ and $N_{R}$ denote the number of initial condition, boundary condition and residual points respectively. 

The total loss function for training is a weighted sum of the three losses:
\begin{equation}
	L=\lambda_{BC}L_{BC}+\lambda_{IC}L_{IC}+\lambda_RL_R
\end{equation}
where the parameters $\lambda_{IC}$, $\lambda_{BC}$ and $\lambda_{R}$  are weight coefficients to balance the contributions of each loss. These coefficients can be predetermined empirically or treated as hyperparameters and adjusted adaptively during training. 

\subsection{Multi-layer Perceptron and Kolmogorov-Arnold Network}\label{subsec22}

In PINNs, the solution $u(x,\theta)$ is typically modeled by a multi-layer perceptron (MLP) composed of a fixed number of linear layers, each followed by a nonlinear activation function. The parameters are represented as $\theta=\left\{W^{(l)},b^{(l)}\right\}_{l=1}^L$where $w(l)$ and $b(l)$ are the weight matrix and bias of the $l$-th layer, respectively. An MLP with $L$ layers and activation function $\mathbf{\sigma}$ can be expressed as:
\begin{equation}
	u(x,\theta)=[\Phi^1\Phi^2\cdotp\cdotp\cdotp\Phi^L](x)\label{EQ6}
\end{equation}
where:
\begin{equation}\Phi^{(l)}(x^{(l)})=\sigma(W^lx^l+b^l)\end{equation}
Thus, the total number of parameters for the MLP ($|\theta|$) is as follows:
\begin{equation}|\theta|_{MLP}=H[l+(n_l-1)H+0]{\sim}\mathcal{O}(n_lH^2)\end{equation}

where $I$ and $O$ represent the number of inputs and outputs, $n_{l}$ is the number of hidden layers, and $H$ denotes the number of neurons in each hidden layer.

Unlike the universal approximation theorem, which underpins the MLP, the Kolmogorov-Arnold representation theorem establishes that any multivariate continuous function $\mathrm{f}$ on a bounded domain can be represented as:
\begin{equation}f(x_1,x_2\cdots,x_n)=\sum_{q=1}^{2n+1}\Phi_q(\sum_{p=1}^n\phi_{q,p}(x_p))\end{equation}

where $\Phi_{q,p}$ is a mapping from $[0,1]{\to}\mathbb{R}$ , $\Phi_{q}$ is a mapping from $\mathbb{R}{\rightarrow}\mathbb{R}$. However, the one-dimensional functions may exhibit non-smooth or fractal-like characteristics, making them impractical for direct learning. To address this, Liu et al \cite{b36}. introduced a neural network design inspired by this theorem, ensuring all learnable functions are univariate and parameterized as B-splines, enhancing flexibility and learnability.

In the original KAN implementation, the univariate activation function is defined as:
\begin{equation}\phi(x)=c_rr(x)+c_BB(x)\end{equation}
where:
\begin{equation}r(x)=\frac{x}{1+exp(-x)}\end{equation}
\begin{equation}B(x)=\sum_{i=1}^{G+k}c_iB_i(x)\end{equation}
where $r(x)$ is the basis function, $B(x)$ is the $k$-th order $B$-spline defined on a grid with $G$ intervals. For a given grid $g$ and $B$-spline order $k$, a unique set of spline basis functions $\{B_i\}_{i=1}^{G+k}$, is defined, with the parameters $C_{r}, C_{B}$ and $\{c_i\}_{i=1}^{G+k}$ being trainable. These features make the activation functions in KAN adaptable rather than fixed.
In KAN, while Equation \eqref{EQ6} remains valid, the nonlinear activations in the linear combinations are replaced by:
\begin{equation}\begin{gathered}
		\Phi^{(l)}\left(x^{(l)}\right)=
		\begin{pmatrix}
			\phi_{l,1,1}(\cdot) & \phi_{l,1,2}(\cdot) & ... & \phi_{l,1,m_l}(\cdot) \\
			\phi_{l,2,1}(\cdot) & \phi_{l,2,2}(\cdot) & ... & \phi_{l,2,m_l}(\cdot) \\
			\vdots & & \ddots & \vdots \\
			\phi_{l,m_{l+1},1}(\cdot) & \phi_{l,m_{l+1},2}(\cdot) & ... & \phi_{l,m_{l+1},m_l}(\cdot)
		\end{pmatrix}x^{(l)}
\end{gathered}\end{equation}
where $m_{l}$ is the number of input nodes in the $l$-th layer, and $\phi_{l,i,j}$ is the univariate activation function in the $l$-th layer that connects the $i$-th input node in the network computational graph to the $j$-th output node. Essentially, in KAN, the concept of layers does not revolve around a set of nodes but rather refers to edges, which is where the activation functions are located. For a layer with $m_{l}$ input nodes and $m_{l+1}$ output nodes, the number of univariate activation functions is $m_l\cdot m_{l+1}$. The output of a general KAN network for an input vector $x$ is given by:
\begin{equation}KAN(x)=(\Phi^1\Phi^2\cdots\Phi^l)x\end{equation}
The total parameter count in KAN is calculated as:
\begin{equation}|\theta|_{KAN}=H[l+(n_l-1)H(k+g)+O]{\sim}O(n_lH^2(k+g))\end{equation}
where $I$ and $O$ represent the number of inputs and outputs, $n_{l}$ is the number of hidden layers, $H$ is the number of neurons in each hidden layer, $g$ is the grid size, and $k$ is the polynomial order.

\begin{figure}[h]
	\centering
	\includegraphics[width=1\textwidth]{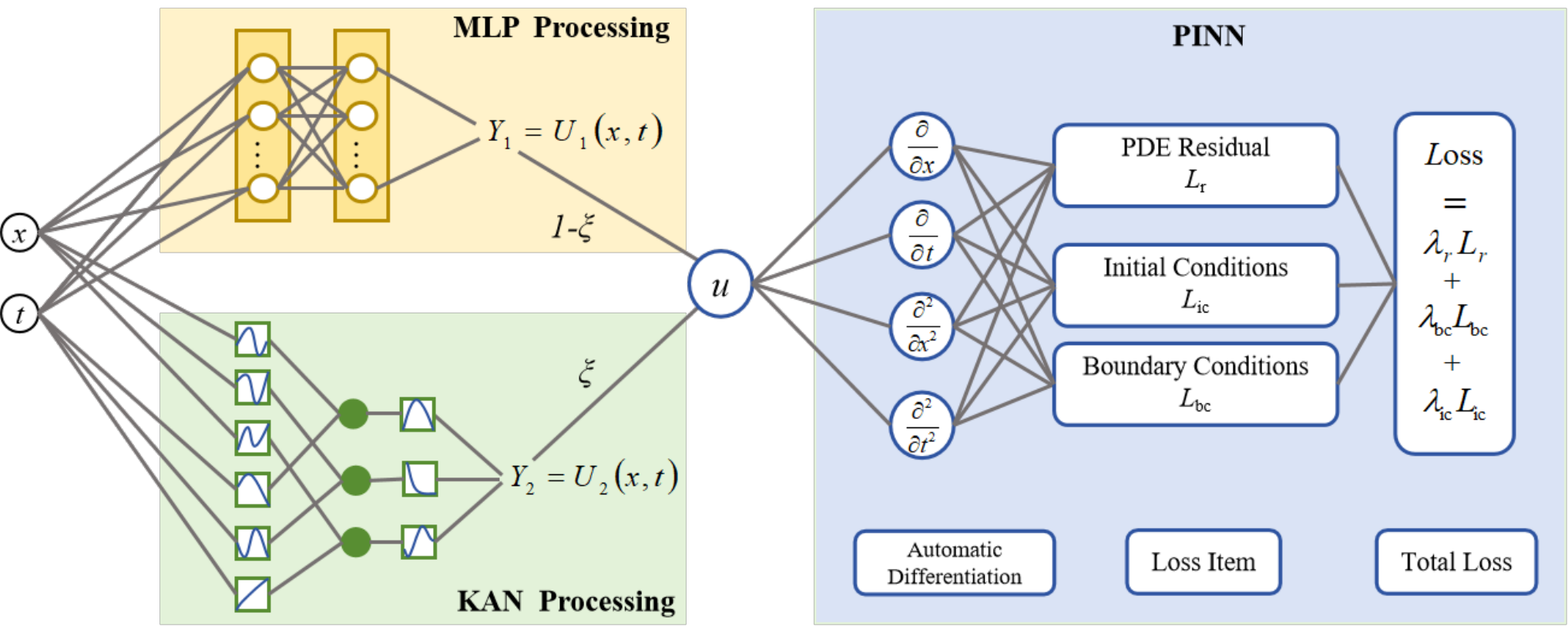}
	
	\caption{Hybrid Parallel KAN-MLP PINN architecture. On the left side of the figure is the neural network component, while the right side represents the physics-informed component. The neural network part receives two input variables, the spatial variable $x$ and the temporal variable $t$, which are simultaneously fed into two parallel networks: MLP and KAN. Each network processes the inputs through their respective hidden layers, generating outputs $Y_{1}$ and $Y_{2}$. These outputs are then combined using an adjustable network weight factor $\xi$, resulting in the final network output $u=\xi\cdot Y_2+(1-\xi)\cdot Y_1$. This output is subsequently passed to the following physics-informed component to generate the corresponding loss function term.}
	\label{figure1}
\end{figure}

\subsection{HPKM-PINN}\label{subsec23}
The HPKM-PINN architecture integrates the strengths of two distinct neural network types—MLP and KAN—in a parallel structure. This combination leverages the complementary capabilities of each network to enhance the performance of PINNs in solving complex mathematical problems, such as function fitting and PDEs. The specific parallel structure is illustrated in Fig. \ref{figure1}.

The HPKM-PINN model employs a parallel-output structure where the outputs of the MLP and KAN are fused proportionally to form the final prediction. Each network operates independently on the same input variables, processing the data through their respective architectures. The resulting outputs are combined using a scaling factor, denoted by $\xi$, which controls the relative contributions of the MLP and KAN networks to the total output. The combined output $u(x)$ is defined as:
\begin{equation}u(x)=\xi\cdot u_{KAN}(x)+(1-\xi)\cdot u_{MLP}(x)\end{equation}
\section{Numerical experiments}\label{sec3}
In this section, we present numerical experiments to assess the performance of the proposed HPKM-PINN in comparison with conventional methods for function fitting and PDE solving. The experiments are designed to ensure a fair comparison, with all methods employing the same optimization strategy and learning rate settings for both the KAN and MLP components unless otherwise stated. The primary distinction among these methods lies in their respective neural network architectures.

The performance of each model is evaluated using the normalized $L_{2}$ error metric, which measures the deviation of the predicted solution from the reference solution. The $L_{2}$ error is defined as:
\begin{equation}L_2=\frac{\sqrt{\sum_{i=1}^N\left|u(x_i,t_i)-\widehat{u}(x_i,t_i)\right|^2}}{\sqrt{\Sigma_{i=1}^N\left|u(x_i,t_i)\right|^2}}\end{equation}
where $u$ represents the reference solution and $\hat{u}$ denotes the predictions made by the neural network.

\subsection{Approximation of High-Low Frequency Mixed Functions}\label{subsec31}
To evaluate the approximation capabilities of KAN, MLP, and the proposed HPKM, we selected a function that exhibits discontinuities, and incorporates both high- and low-frequency components. This function was chosen specifically to assess the robustness of the hybrid model in mitigating spectral bias \cite{b32,b46}, a phenomenon commonly observed in neural networks, as discussed in. The function is defined as:
\begin{equation}y(x)=
	\begin{cases}
		5+\sum_{k=0}^3\sin\left((k+1)\pi x\right), & x<0 \\
		\sin(2\pi x)+0.5\sin(25\pi x), & x\geq0 
	\end{cases}
\end{equation}
A uniform distribution of 500 points was used over the interval $[-3,3]$. The MLP structure had a structure of $[1,100,100,100,100,1]$, while the KAN used a structure of $[1,5,5,5,1]$ with a grid size of 5. All models employed the Adam optimizer with a learning rate of 0.01 for consistency.

\begin{figure}[h]
	\centering
	\includegraphics[width=1\textwidth]{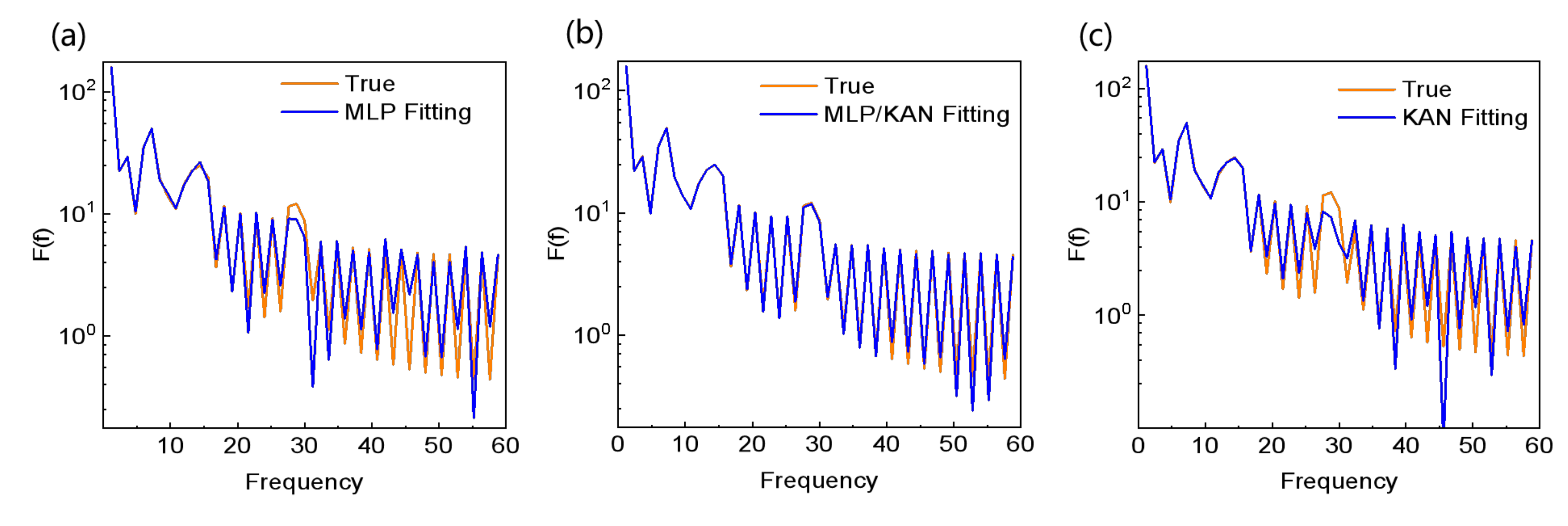}
	
	\caption{Comparison of the Fourier spectra of the reference function, and the approximated functions obtained using (a) KAN Ratio $\xi$ = 0 (MLP), (b) KAN Ratio $\xi$ = 0.9 (hybrid structure), and (c) KAN Ratio $\xi$ = 1 (KAN).}
	\label{figure2}
\end{figure}

Fig. \ref{figure2}(a)-(c) present the Fourier spectra of the reference function along with the approximated spectra generated by the three architectures with different network weight factors ($\xi$ = 0 for MLP, $\xi$ = 0.9 for the HPKM, and $\xi$ = 1 for KAN). The results clearly demonstrate that the MLP model, with $\xi$ = 0, fails to capture the high-frequency components of the function, leading to an incomplete representation of the function's spectral characteristics. Similarly, the KAN model with $\xi$ = 1 also struggles to accurately model the full spectrum of the function, resulting in a poor approximation. These observations highlight the limitations of both the MLP, and KAN architectures when used in isolation for such mixed-frequency function approximation tasks.

\begin{figure}[h]
	\centering
	\includegraphics[width=0.9\textwidth]{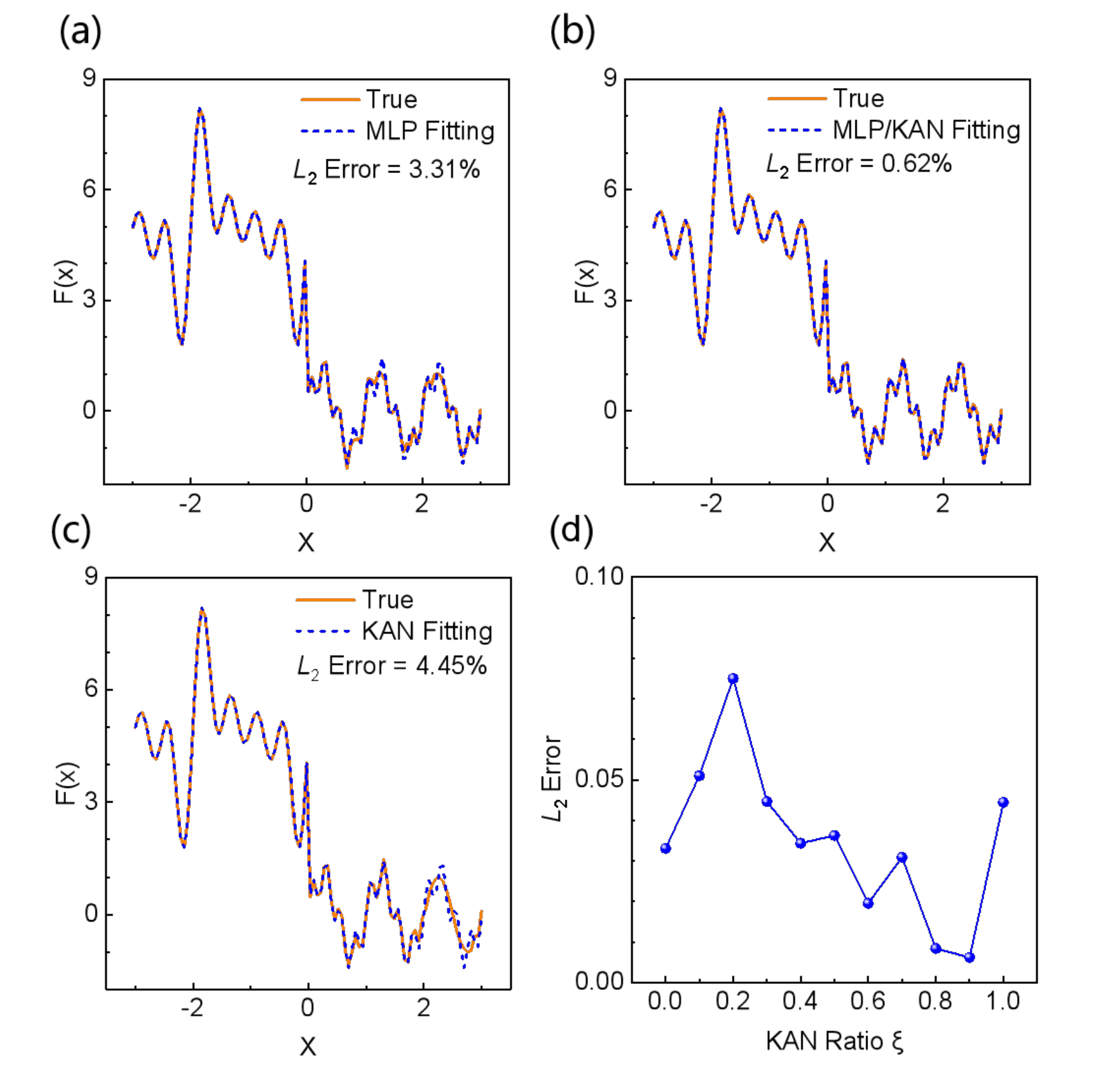}
	
	\caption{Comparison of the predicted solutions and reference solutions for Equation (14) using (a) KAN Ratio $\xi$ = 0 (MLP), (b) KAN Ratio $\xi$ = 0.9 (hybrid structure), and (c) KAN Ratio $\xi$ = 1(KAN). (d) Relationship between the $L_{2}$ error and the network weight factor.}
	\label{figure3}
\end{figure}

In contrast, the HPKM model with $\xi$ = 0.9 significantly improves the approximation. Fig. \ref{figure3}(a)-(c) display the function approximation results for each architecture, showing relative $L_{2}$ errors of 3.31\%, 0.62\%, and 4.45\% for the MLP, HPKM, and KAN models, respectively. To further investigate the impact of the network weight factor on the model's performance, we conducted a systematic evaluation by varying $\xi$ from 0 to 1 in increments of 0.1. As shown in Fig. \ref{figure3}(d), the $L_{2}$ errors fluctuated significantly across different weight factors, with the optimal performance achieved at $\xi$ = 0.9. This confirms that the hybrid architecture, by combining the strengths of both KAN and MLP, is particularly effective in approximating functions with mixed high- and low-frequency components.

\begin{figure}[h]
	\centering
	\includegraphics[width=1\textwidth]{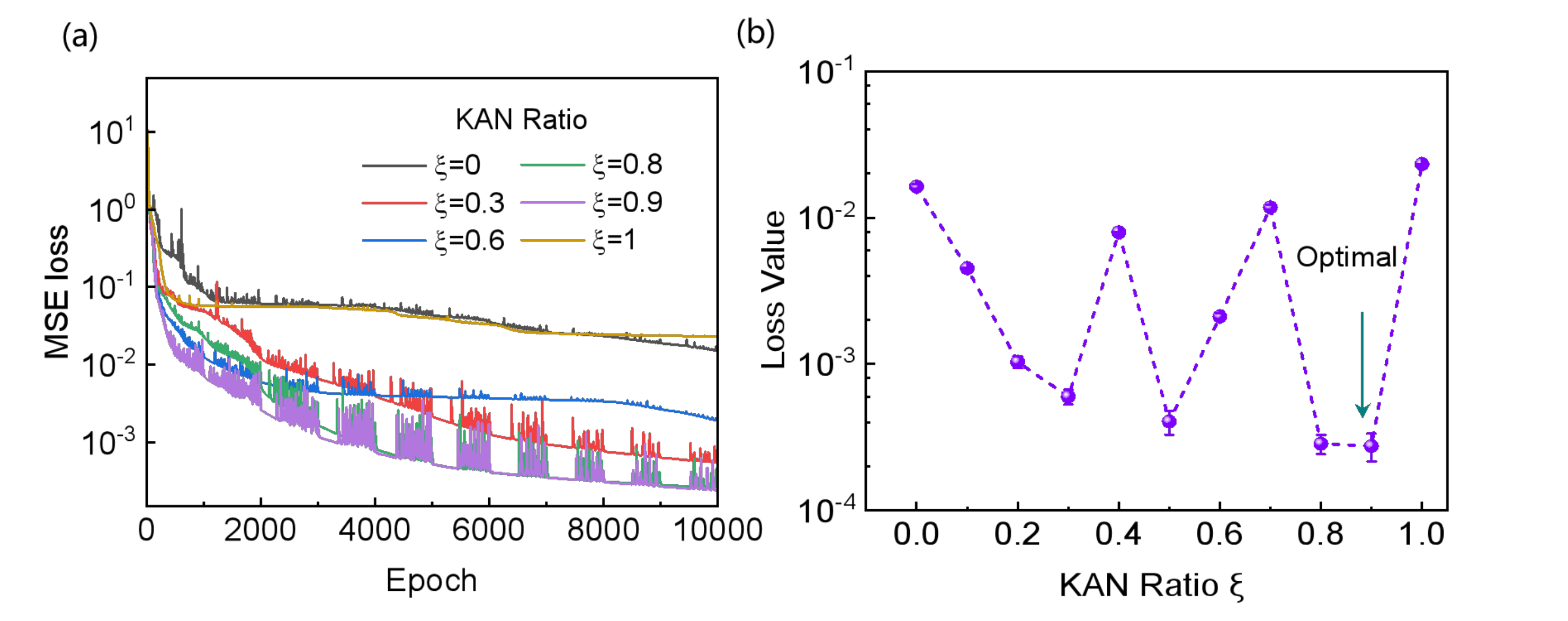}
	
	\caption{(a) Training progression of MSE loss across epochs for different KAN ratios. (b) Final convergence loss values versus KAN ratio, with error bars reflecting stability during the final 500 training epochs.}
	\label{figure4}
\end{figure}

Additionally, we observed that the hybrid model outperformed both the MLP and KAN models in terms of both fitting accuracy, and the ability to capture the function's full spectral range. The HPKM exhibited superior capability in managing the function’s mixed-frequency characteristics, leading to accelerated convergence and enhanced accuracy across both low- and high-frequency components. Fig. \ref{figure4} illustrates the convergence curves of the loss functions for all three architectures, each trained for 10,000 iterations. The results demonstrate that the hybrid parallel architecture converges most rapidly, attaining the minimum loss value sooner than the MLP and KAN architectures, with optimal performance observed at $\xi$ = 0.9. This finding further substantiates the effectiveness of the hybrid approach in overcoming the challenges associated with complex mixed-frequency function approximation.

\subsection{PDE Solving}\label{subsec32}
The primary focus of this work is to demonstrate the capability of the HPKM-PINN in solving PDEs. In this section, we conduct a series of numerical experiments on four distinct types of PDEs to validate the effectiveness and robustness of the proposed HPKM-PINN parallel architecture in practical problem-solving scenarios. Through these experiments, we aim to showcase how the hybrid model compares to traditional approaches, particularly in terms of accuracy, convergence speed, and overall efficiency.
\subsubsection{Poisson equation}\label{subsubsec321}
We begin by investigating the well-known one-dimensional Poisson equation \cite{b47}, which is commonly used in computational physics and engineering problems. The equation is given by:

\begin{equation}-\frac{\mathrm{d}^2u}{\mathrm{d}x^2}=\mathrm{f}(x)\quad x\epsilon[-L,L]\end{equation}
where the boundary conditions are defined as $u(-L){=}u(L){=}0$. We set $L=2\sqrt{\pi}$, and the exact solution is given by $u(x)=\sin(x^2)$.
To generate the training data, we use Sobol sequences, which provide a high-quality distribution of points across the domain. In total, we generate 2000 residual points within the interval $[-L,L]$ to compute the residual of the PDE, and 500 boundary condition data samples for the prescribed boundary conditions at $x=\pm L$.

The MLP network used for this experiment has a simple architecture of $[1,20,20,1]$, with two hidden layers of 20 units each. For the KAN, we use a slightly more complex architecture $[1,30,30,1]$, with a grid size of 5 and a polynomial order of 3, to better capture the complexity of the solution. Both networks are trained using the Adam optimizer with a learning rate of 0.001, for a total of 15,000 iterations. Aside from the differing network structures, all other parameters are kept constant across the models.

\begin{figure}[H]
	\centering
	\includegraphics[width=0.9\textwidth]{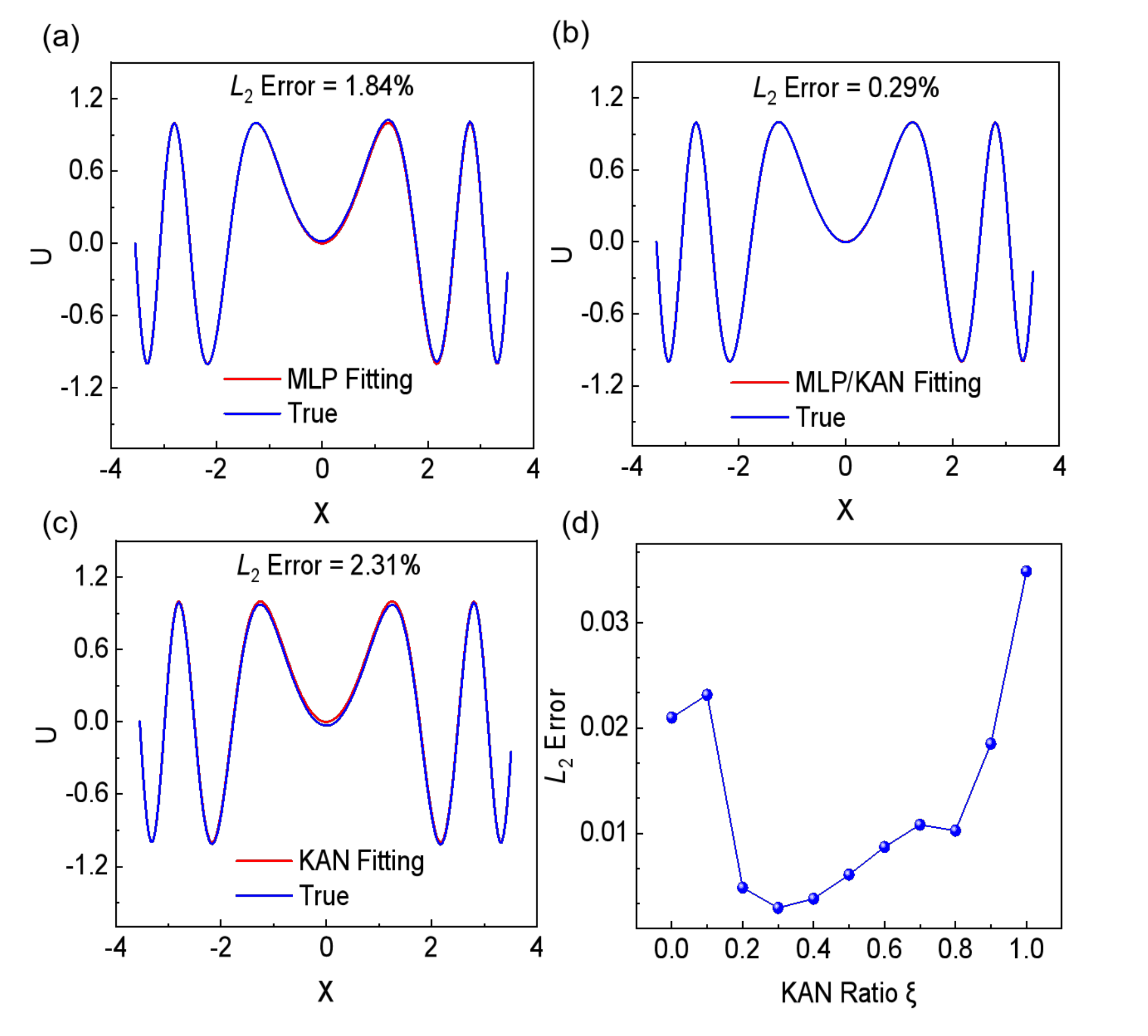}
	
	\caption{Three PINN models for solving the Poisson equation. (a) MLP-based PINN, with a prediction error of 1.84\%. (b) The proposed HPKM-PINN with a network weight factor $\xi$ = 0.3, yielding a prediction error of 0.29\%. (c) KAN-based PIKAN, with a prediction error of 2.31\%. (d) Relationship between $L_{2}$ error and network weight factors.}
	\label{figure5}
\end{figure}

After training, we evaluate the performance of each model by comparing the predicted solution to the exact solution $u(x)=\sin(x^2)$ using the relative $L_{2}$ error. Theresults are shown in Fig. \ref{figure5}(a)-(c), where the approximations for the different network weight factors ($\xi$ = 0 for PINN, $\xi$ = 0.3 for HPKM-PINN, and $\xi$ = 1 for PIKAN) are presented. The corresponding relative $L_{2}$ errors are 1.84\%, 0.29\% and 2.31\%, respectively, indicating that the HPKM-PINN with $\xi$ = 0.3 outperforms both the standard PINN and KAN models in terms of fitting accuracy.

In Fig. \ref{figure5}(d), we provide a more detailed analysis of the effect of varying the network weight factor $\xi$ on the $L_{2}$ error, showing how the error fluctuates across different values of $\xi$. The results demonstrate that the HPKM is most effective when $\xi$ is set to 0.3, which balances the strengths of both KAN and MLP for this specific problem.

\begin{figure}[H]
	\centering
	\includegraphics[width=1\textwidth]{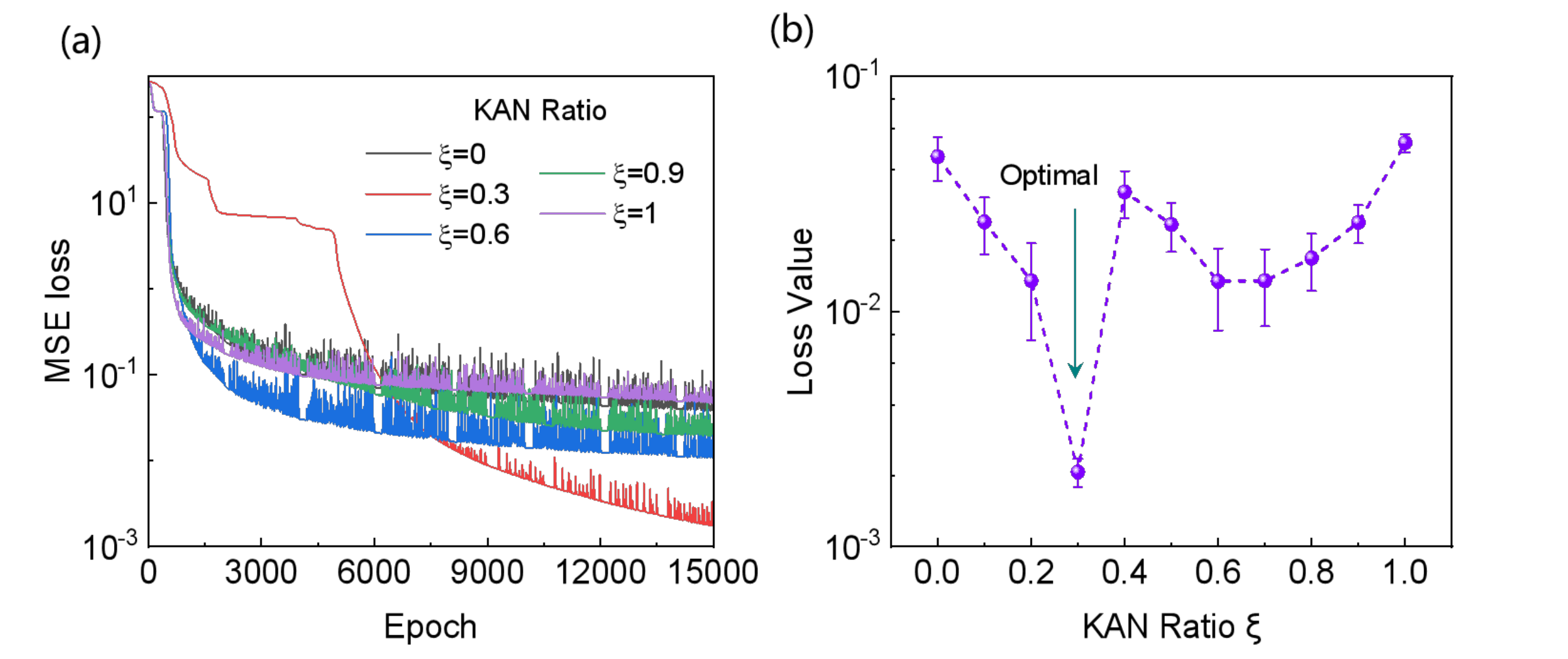}
	
	\caption{(a) Training progression of MSE loss across epochs for different KAN ratios. (b) Final convergence loss values versus KAN ratio, with error bars reflecting stability during the final 500 training epochs.}
	\label{figure6}
\end{figure}

Additionally, we analyze the convergence of the loss function for the three architectures. Fig. \ref{figure6} shows the convergence behavior for the PINN, HPKM-PINN, and PIKAN models during the 15,000 iterations of training. The hybrid parallel model (HPKM-PINN with $\xi$ = 0.3) exhibits the fastest convergence, reaching a minimum loss value more quickly than either the PINN or PIKAN models. This further reinforces the advantage of the HPKM-PINN architecture in terms of both speed, and accuracy when solving PDEs like the Poisson equation.

The promising results of this experiment demonstrate the potential of the HPKM-PINN framework to efficiently and accurately solve a broad class of PDEs. In the next sections, we will extend these findings by applying the HPKM-PINN architecture to additional PDEs to further validate its generalizability and robustness.

\subsubsection{Advection equation}\label{subsubsec322}
Advection, a fundamental process in atmospheric motion, plays a crucial role in the transport of physical quantities such as heat, pollutants and momentum. The governing equation of advection, which includes an advection term, is typically expressed as a first-order partial differential equation \cite{b48}. In this study, we consider the advection equation in one dimension as follows:
\begin{equation}\frac{\partial u}{\partial t}+\frac{\partial u}{\partial x}=0\quad x\in[0,1],t\in[0,0.5]\end{equation}
with the initial/boundary conditions defined as:
\begin{equation}\begin{aligned}
		& u(0,x)=2\sin(\pi x) \\
		& u(t,0)=-2\sin(\pi t),\quad u(t,1)=2\sin(\pi t)
\end{aligned}\end{equation}
The exact solution for the advection equation in this example is given by $u(x,t)=2\sin\left(\pi(x-t)\right)$.

To solve this advection equation using neural networks, we generate a dataset comprising 2,000 residual points, 500 initial condition sample points and 500 boundary condition sample points. To further enhance the training efficiency and speed of convergence, we implement a learning rate scheduler, StepLR, which reduces the learning rate to 75\% of its original value every 1,000 epochs ($\gamma$ = 0.75). This dynamic adjustment helps the model converge quickly in the early stages of training, and fine-tune the learning process as it approaches an optimal solution. This scheduling strategy allows the model to better capture the physical phenomena associated with the advection equation by allowing a more nuanced optimization as training progresses.

\begin{figure}[H]
	\centering
	\includegraphics[width=1\textwidth]{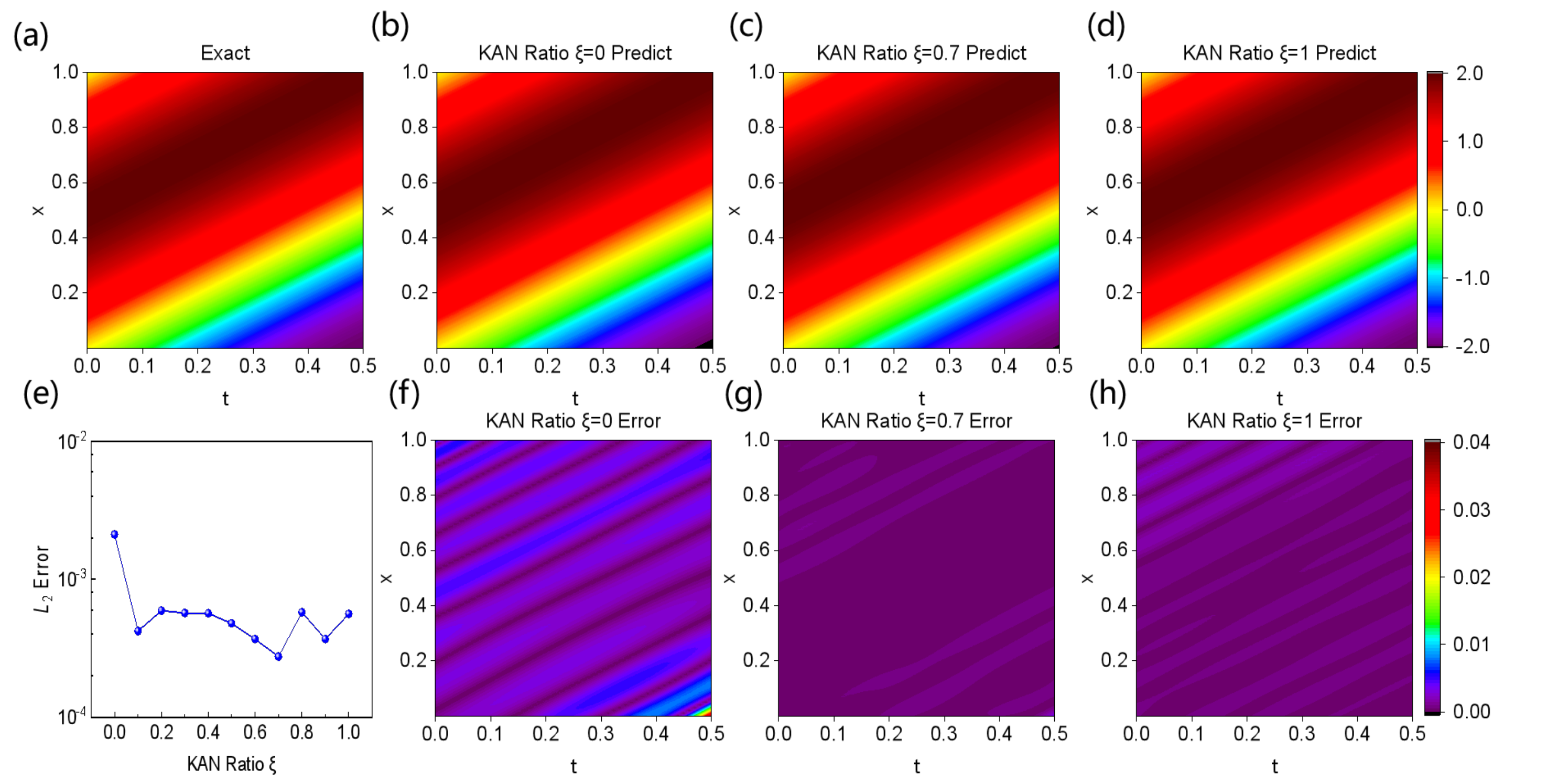}
	
	\caption{presents the solution to the Advection equation in the spatiotemporal domain, where Subfigure (a) shows the reference solution. Solutions obtained from (b) PINN, (c) HPKM-PINN and (d) PIKAN architectures are also displayed. Subfigure (e) illustrates the relationship between the $L_{2}$ error and the network weight factors. Subfigures (f), (g), and (h) represent the absolute pointwise errors between the reference solution and the solutions obtained from PINN, HPKM-PINN and PIKAN, respectively.}
	\label{figure7}
\end{figure}

After training, the performance of the trained models is evaluated by calculating the absolute pointwise error between the predicted solution and the exact solution over the test set. The results, presented in Fig. \ref{figure7},  illustrate the predictive accuracy of the model across the entire computational domain. Specifically, we evaluate the relative $L_{2}$ errors for different architectures with network weight factors $\xi$ = 0 (PINN), $\xi$ = 0.7 (HPKM-PINN), and $\xi$ = 1 (PIKAN), yielding values of 0.21\%, 0.028\%, and 0.056\%, respectively. These results demonstrate that the HPKM-PINN model outperforms both the standard PINN and the KAN-based PIKAN model in terms of accuracy, with the hybrid model showing significantly lower errors across the domain.

\begin{figure}[h]
	\centering
	\includegraphics[width=1\textwidth]{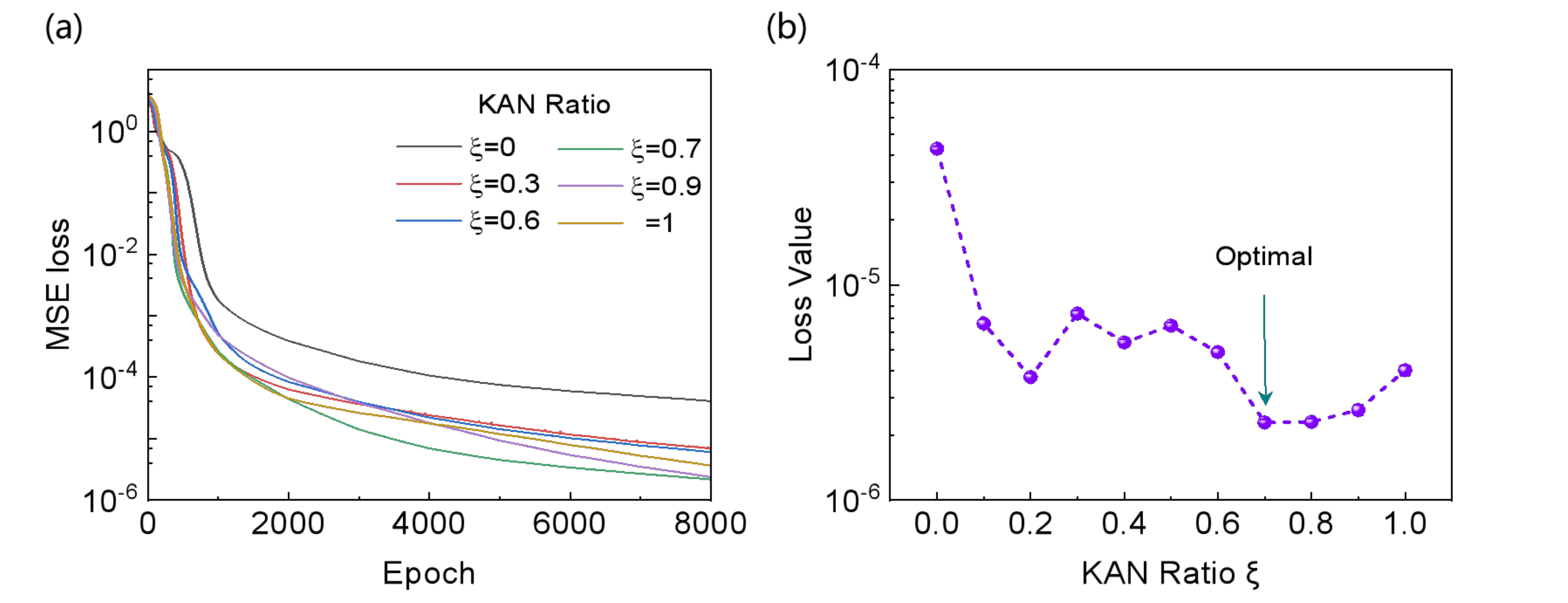}
	
	\caption{(a) Training progression of MSE loss across epochs for different KAN ratios. (b) Final convergence loss values versus KAN ratio, with error bars reflecting stability during the final 500 training epochs.}
	\label{figure8}
\end{figure}

As shown in Fig. \ref{figure8}, the training process exhibits a characteristic trend in the loss function. Initially, the loss function decreases rapidly, reflecting the model's efficient learning in the early epochs. After this initial phase, the loss function stabilizes, indicating that the model has successfully converged and reached an optimal solution. This behavior further confirms that the HPKM-PINN model is able to efficiently capture the dynamics of the advection process and achieve satisfactory training performance.

The experimental results demonstrate the effectiveness of HPKM-PINN model in solving the advection equation. The HPKM-PINN model outperforms traditional PINN, and PIKAN architectures in terms of both fitting accuracy and convergence speed. It effectively captures the convective phenomena present in the advection process, providing a robust approach for solving partial differential equations with mixed high- and low-frequency characteristics.

\subsubsection{Convention-Diffusion equation}\label{subsubsec323}
The convection-diffusion equation is a fundamental partial differential equation widely used to model a range of physical phenomena, including heat transfer, pollutant dispersion, and fluid flow. This equation describes the combined effects of convection (transport due to a moving fluid) and diffusion (spreading due to molecular motion). The general form of the convection-diffusion equation is given by:
\begin{equation}\frac{\partial u}{\partial t}+c\cdot\frac{\partial u}{\partial x}=\mu\cdot\frac{\partial^2u}{\partial x^2} \quad x\in[-L,L],t\in[0,T]\end{equation}
where $u(x,t)$ represents the quantity of interest (such as concentration or temperature), $c$ is the convection speed, and $\mu$ is the diffusion coefficient. In this study, we consider the following initial and boundary conditions for the problem:
\begin{align}
	&u(0,x)=\frac{0.1}{\sqrt{0.1\cdot\mu}}\cdot exp\left(-\frac{(x+2)^2}{4\cdot0.14\cdot\mu}\right) \notag \\
	&u(t,-L)=u(t,L)=0
\end{align}
where the initial condition represents a Gaussian distribution that is centered at $x=-2$, and the boundary conditions are zero at both ends of the domain. The parameters for this problem are set as follows: $c$ = 4, $\mu$ = 0.05, $L$ = 4 and $T$ = 1, which represent typical values for convective and diffusive processes in practical applications.

\begin{figure}[H]
	\centering
	\includegraphics[width=1\textwidth]{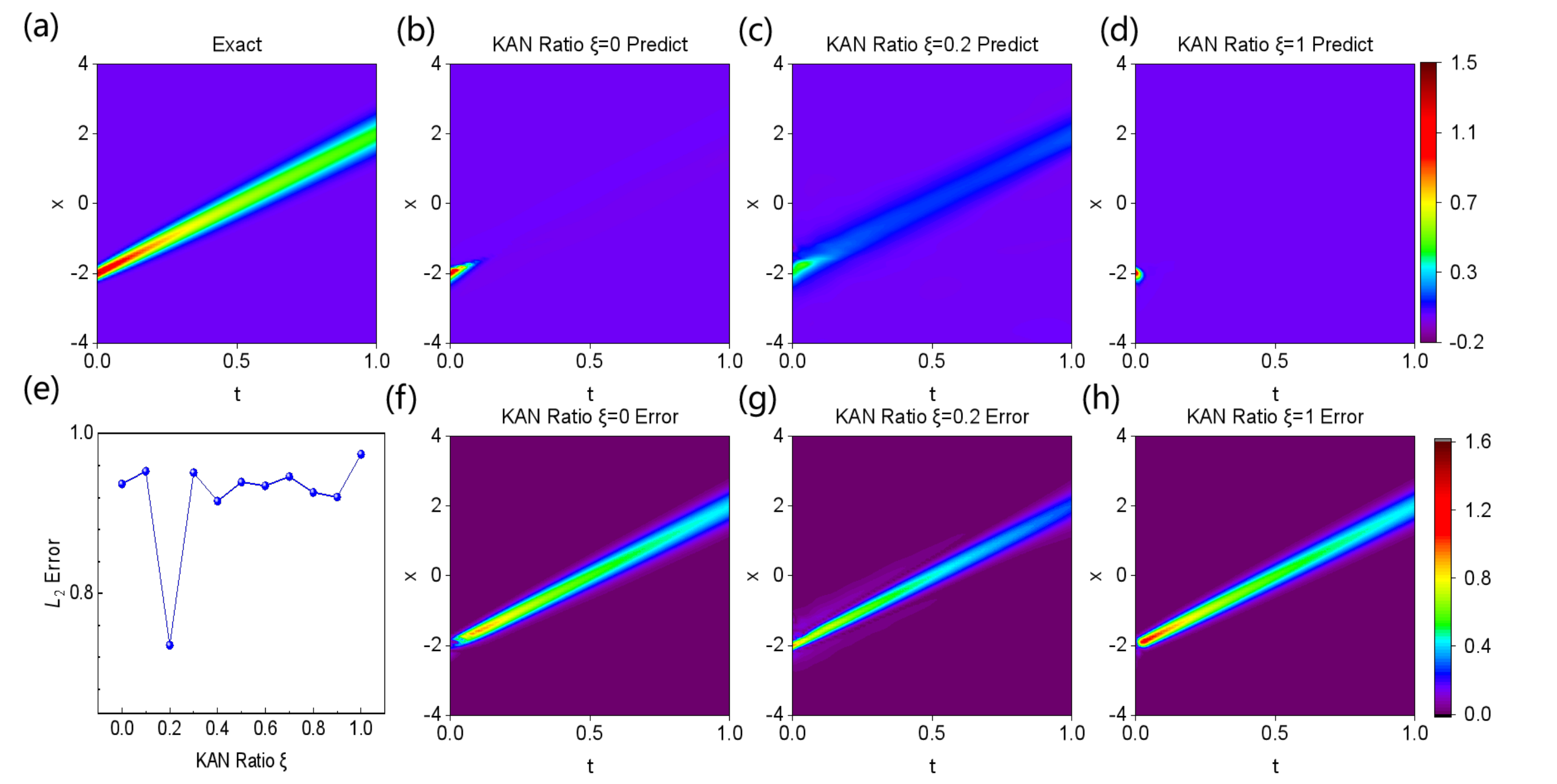}
	
	\caption{presents the solutions to the advection-diffusion equation in the spacetime domain, Subfigure (a) shows the reference solution. Solutions obtained from (b) PINN, (c) HPKM-PINN, and (d) PIKAN architectures are also displayed. Subfigure (e) illustrates the relationship between the $L_{2}$ error and the network weight factors. Subfigures (f), (g), and (h) represent the absolute pointwise errors between the reference solution and the solutions obtained from PINN, HPKM-PINN, and PIKAN, respectively.}
	\label{figure9}
\end{figure}

The model configuration and data generation process are consistent with the setup used for the previous advection equation experiment. Specifically, we generate a set of 2,000 residual points, 500 initial condition points, and 500 boundary condition points for training the neural network models. Fig. \ref{figure9} displays the pointwise error plots for the three models: the PINN with $\xi$ = 0, the HPKM-PINN with $\xi$ = 0.2, and the KAN-based model (PIKAN) with $\xi$ = 1. The relative $L_{2}$ errors for these models are 0.94, 0.74 and 0.97, respectively. These results reveal that the HPKM-PINN model, with its parallel architecture, provides a better fit to the convection-diffusion equation than either the standard PINN or PIKAN models, showing a significant improvement in terms of accuracy.

\begin{figure}[H]
	\centering
	\includegraphics[width=1\textwidth]{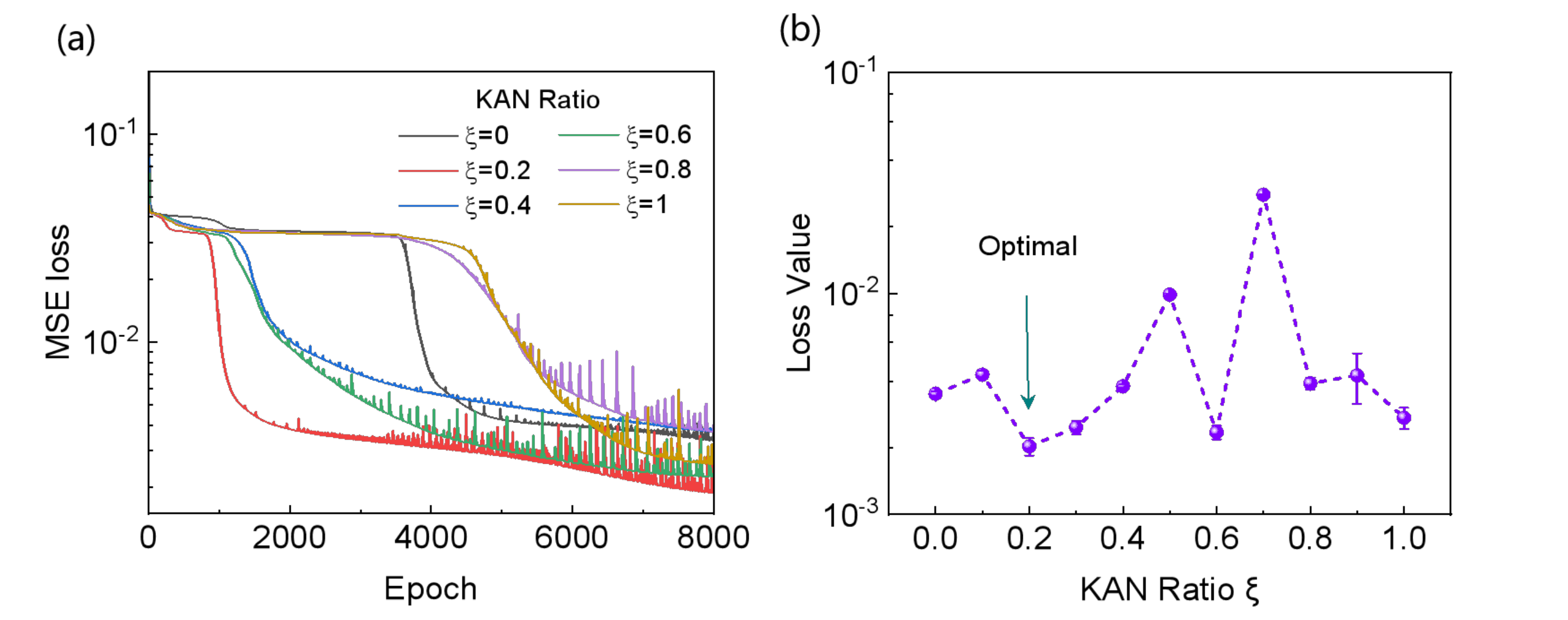}
	
	\caption{(a) Training progression of MSE loss across epochs for different KAN ratios. (b) Final convergence loss values versus KAN ratio, with error bars reflecting stability during the final 500 training epochs.}
	\label{figure10}
\end{figure}

In Fig. \ref{figure10}, we present the convergence behavior of the loss function during the training process. As indicated, both the PINN and PIKAN models exhibit an early stagnation in convergence, where the loss function fails to decrease significantly in the initial iterations. This early stagnation is a known challenge in training deep learning models for complex partial differential equations \cite{b13}. However, the HPKM-PINN model successfully overcomes this issue. As seen in the plot, the loss function decreases rapidly after around 2,000 iterations and continues to decline until it stabilizes at a minimum value, indicating efficient convergence. This enhanced training performance highlights the effectiveness of the hybrid parallel structure in tackling the complexities posed by the convection-diffusion equation.

The experimental results demonstrate the advantages of the HPKM-PINN model in solving the convection-diffusion equation, particularly in the presence of more complex initial conditions. By combining the MLP and KAN architectures in a parallel structure, the HPKM-PINN model not only achieves higher accuracy but also overcomes the convergence issues commonly faced by traditional neural network approaches. The ability to capture both convective and diffusive phenomena with greater efficiency and accuracy makes the hybrid parallel model a promising tool for solving a wide range of physical problems governed by convection-diffusion processes. This study further underscores the potential of hybrid neural network architectures in solving complex partial differential equations with challenging boundary and initial conditions.


\subsubsection{Helmholtz equation}\label{subsubsec324}
In the final example, we examine the Helmholtz equation \cite{b49}, a fundamental elliptic partial differential equation widely used to describe the behavior of electromagnetic waves, acoustics, and other wave phenomena in various physical systems. The equation can be written as:
\begin{equation}
	\frac{\partial^2u}{\partial x^2}+\frac{\partial^2u}{\partial y^2}+k^2u-q(x,y)=0 \quad    x,y\in[-L,L]
\end{equation}
where $u(x,y)$ represents the solution of the equation, $k$ is the wave number, and $q(x,y)$ is a forcing term that introduces inhomogeneity into the system. For the sake of simplicity, we set the wave number $k$ = 1.0 and the domain length $L$ = 1.0. The forcing term $q(x,y)$ is defined as:
\begin{equation}\begin{aligned}
		q(x,y)  =&-(a_1\pi)^2\sin(a_1\pi x)\sin(a_2\pi y) \\
		& -(a_2\pi)^2\sin(a_1\pi x)\sin(a_2\pi y) \\
		& +ksin(a_1\pi x)\sin(a_2\pi y)
\end{aligned}\end{equation}
where $a_{1}$ = 1 and $a_{2}$ = 4 are constants that define the spatial frequencies in the $x$ and $y$ directions. This leads to the analytical solution for the Helmholtz equation: $u(x,y)=\sin(a_1\pi x)\sin(a_2\pi y)$ (with $a_{1}$ = 1 and $a_{2}$ = 4). The boundary conditions for this problem are expressed as:
\begin{equation}u(-1,y)=u(1,y)=u(x,-1)=u(x,1)=0\end{equation}
In solving the Helmholtz equation, we adopted the same experimental settings, and parameter configurations as those used for the previous partial differential equations, ensuring a consistent evaluation framework. This includes the architecture of the neural network, the method for generating training data, and other critical settings such as optimizer choices, and learning rates. Maintaining consistency across these parameters allows for a direct comparison of results, ensuring that any differences in performance can be attributed to the specific nature of the equation being solved rather than changes in the experimental setup. By using a uniform approach, we are able to systematically assess how well the model adapts to and solves various types of partial differential equations.

\begin{figure}[H]
	\centering
	\includegraphics[width=1\textwidth]{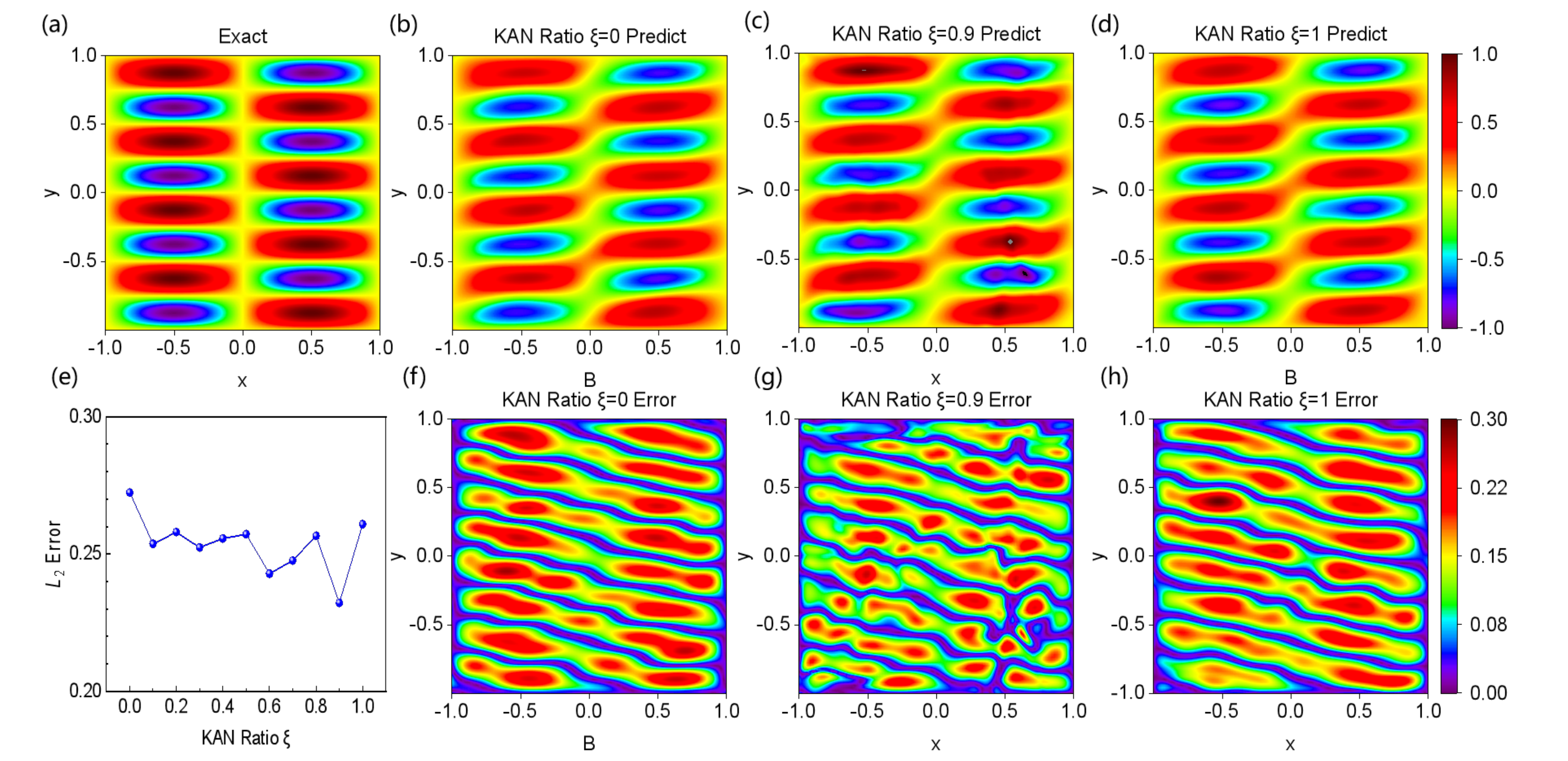}
	
	\caption{The solution of the Helmholtz equation in the spacetime domain is presented. Subfigure (a) shows the reference solution. Solutions obtained from (b) PINN, (c) HPKM-PINN, and (d) PIKAN architectures are also displayed. Subfigure (e) illustrates the relationship between the $L_{2}$ error and the network weight factors. Subfigures (f), (g), and (h) represent the absolute pointwise errors between the reference solution and the solutions obtained from PINN, HPKM-PINN and PIKAN, respectively.}
	\label{figure11}
\end{figure}

The results from solving the Helmholtz equation are summarized in terms of relative $L_{2}$ errors for the three models: the PINN with network weight factor $\xi$ = 0, the HPKM-PINN with $\xi$ = 0.9 and the PIKAN with $\xi$ = 1. The relative $L_{2}$ errors for the respective models are: PINN ($\xi$ = 0): 0.27, HPKM-PINN ($\xi$ = 0.9): 0.23, PIKAN ($\xi$ = 1): 0.26,
These results demonstrate that while all models achieve relatively low errors, there is only a slight difference in performance between the models. The hybrid HPKM-PINN model performs slightly better in terms of error reduction when compared to the PINN and PIKAN models. However, as shown in Fig. \ref{figure11}, the improvement in predictive accuracy is not dramatic. While some localized improvements in the solution can be seen, the overall accuracy remains similar across all models. This suggests that for the specific configuration of the Helmholtz equation studied here, the use of a hybrid parallel model may not yield substantial improvements over more traditional approaches, particularly when the problem's structure is relatively simple and well-behaved.

The solution of the Helmholtz equation highlights the efficiency of deep learning models, particularly hybrid parallel networks, in solving elliptic partial differential equations. Although the HPKM-PINN model provides some localized improvements in terms of accuracy, the performance of all models remains comparable for this particular example. This indicates that for simpler physical systems described by elliptic equations, traditional approaches, such as PINNs or KANs, can achieve satisfactory results without the need for complex hybrid structures. However, as more complex scenarios are considered—such as those with more intricate boundary conditions or forcing terms—the benefits of hybrid parallel models like HPKM-PINN are expected to become more pronounced. This experiment underscores the importance of evaluating model performance across a variety of equations to understand where hybrid architectures may offer the most significant advantages in solving real-world problems.

\begin{figure}[H]
	\centering
	\includegraphics[width=1\textwidth]{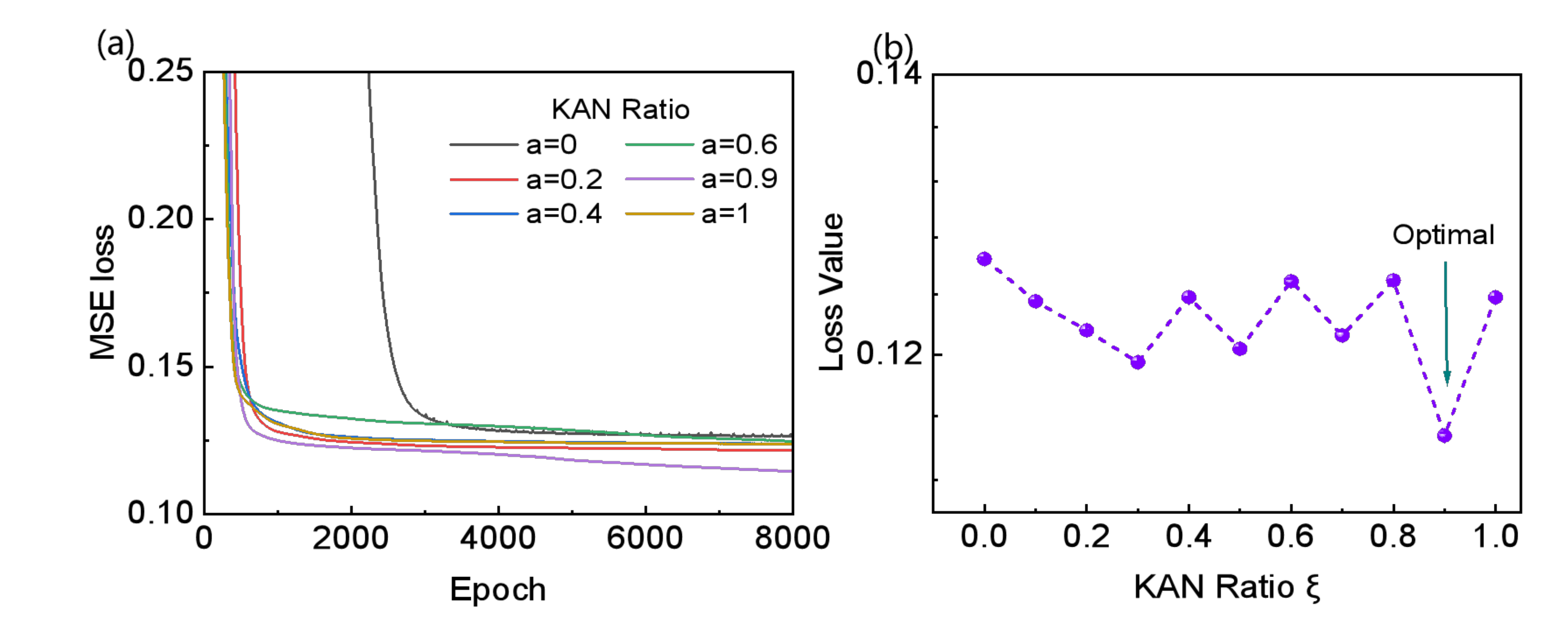}
	
	\caption{(a) Training progression of MSE loss across epochs for different KAN ratios. (b) Final convergence loss values versus KAN ratio, with error bars reflecting stability during the final 500 training epochs.}
	\label{figure12}
\end{figure}

\subsection{Robustness and Limitations}\label{subsec33}
To further assess the robustness of the proposed HPKM-PINN model, we introduced Gaussian noise to the test set. This noise was applied to simulate real-world conditions where data may be contaminated or imperfect. We generated noise with varying intensities within this range and systematically evaluated the model’s performance under noisy conditions. For each noise level, we computed the $L_{2}$ errors to quantify the model’s prediction accuracy as noise intensity increased.

\begin{figure}[H]
	\centering
	\includegraphics[width=0.9\textwidth]{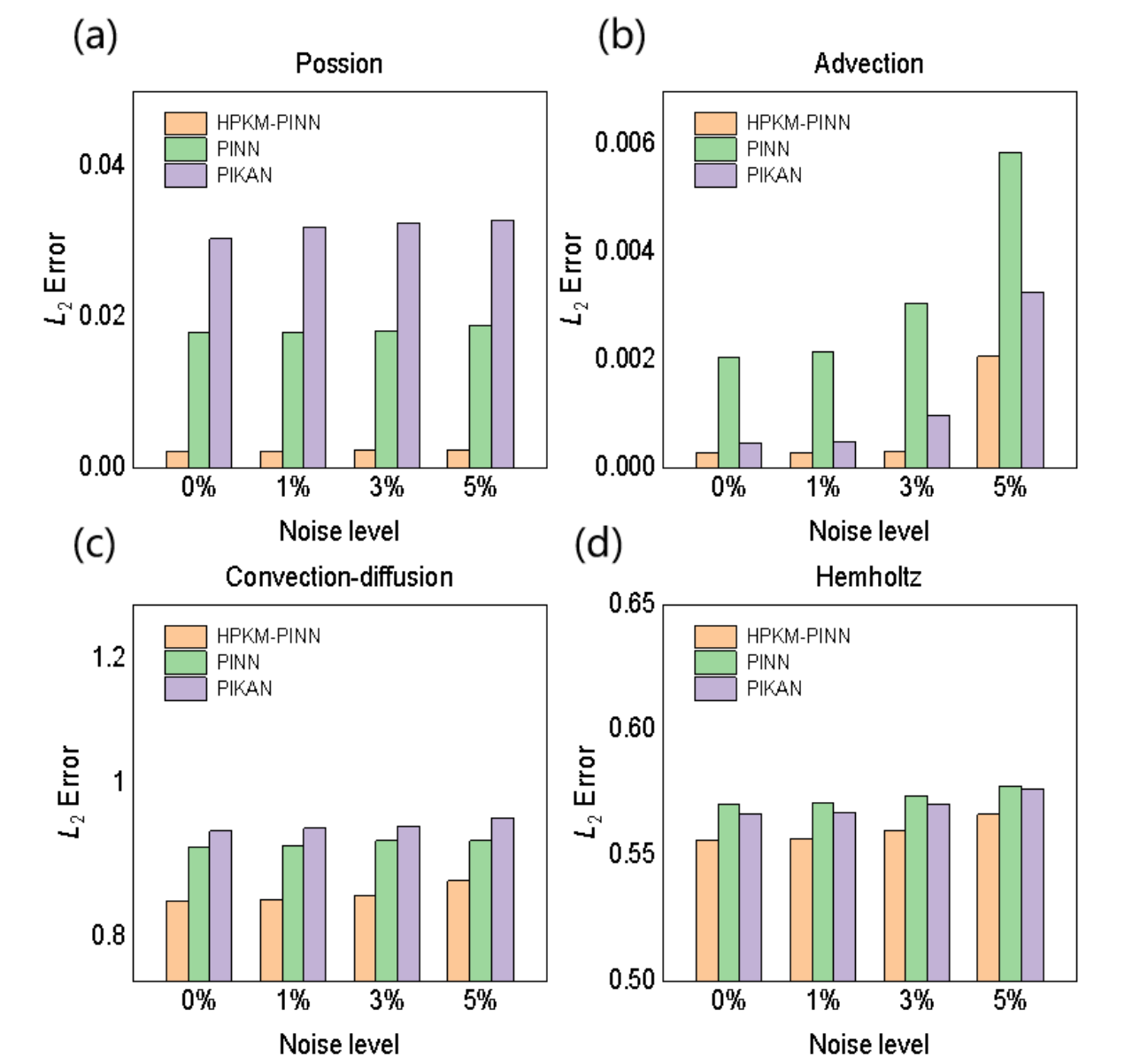}
	
	\caption{A comparison of the $L_{2}$ errors of the three models when trained on clean data but tested with noisy input data in solving (a) Poisson (b) Advection (c) Convection-Diffusion and (d) Helmholtz equations.}
	\label{figure13}
\end{figure}

The results, illustrated in Fig. \ref{figure13}, reveal that HPKM-PINN demonstrates superior stability and performance across different noise levels. Specifically, the optimal mixing ratios for the four test equations—$\xi$ = 0.3, 0.7, 0.2, and 0.9—produced the most stable results. Despite slight fluctuations in error as noise intensity increased, the HPKM-PINN model consistently outperformed both the standard Physics-Informed Neural Network (PINN, $\xi$ = 0) and the standalone Kolmogorov-Arnold Network (PIKAN, $\xi$ = 1). Notably, while the error of the PINN model ($\xi$ = 0) increased dramatically at higher noise levels, HPKM-PINN maintained a more stable error profile, particularly at intermediate noise intensities. 

At extreme noise levels, the PIKAN model exhibited relatively consistent error values, although these values were noticeably larger compared to HPKM-PINN. This indicates that while the KAN component in PIKAN offers robustness to noise, it is still less effective in maintaining low error rates when faced with high-intensity noise. In contrast, HPKM-PINN’s hybrid approach enabled it to maintain a balance between the noise resilience of KAN and the flexibility of MLP, ensuring superior robustness in highly noisy environments.

While the performance gains of HPKM-PINN are clear, there is a trade-off in terms of computational cost. The increased number of parameters introduced by the parallel KAN and MLP structure results in longer training times compared to PINN and PIKAN. As shown in Table \ref{table1}, the network architecture of HPKM-PINN consists of a parallel structure integrating KAN and MLP, which increases model complexity and computational demand. This is an expected outcome, as the parallel design adds additional computational complexity. However, the improved accuracy and robustness demonstrated by HPKM-PINN in noisy and multiscale scenarios suggest that the performance benefits outweigh the additional computational cost for many practical applications. HPKM-PINN offers a highly robust and scalable solution for solving PDEs in noisy and complex environments. The hybrid architecture's ability to adaptively balance the contributions of MLP and KAN ensures stable performance across a wide range of conditions. While the trade-off between computational cost and model performance should be carefully considered, HPKM-PINN represents a promising framework for tackling complex real-world problems where both accuracy and robustness are critical.

\begin{table}[H]
	\centering
	\begin{tabular}{c c c c c c}
		\hline
		Equation                 & Model       & \begin{tabular}[c]{@{}c@{}}Architecture\\of NN\end{tabular}      & Parameters      & Relative error   & \begin{tabular}[c]{@{}c@{}}Time\\(Second)\end{tabular}     \\ 
		\hline
		\multirow{3}{*}{Possion} & PINN        & [1,20,20,1]    & 481     & 1.84E-02                       & 88.07   \\
		& PIKAN        & [1,30,30,1]    & 9600      & 2.31E-02                         & 696.36  \\
		& HPKM-PINN    &                & 10481       & 2.91E-03 (\textbf{$\xi$=0.3})       & 858.73  \\ 
		\hline
		\multirow{3}{*}{Advection} & PINN       & [2,20,20,20,1] & 921     & 2.11E-03                      & 51.53   \\
		& PIKAN        & [2,5,5,1]      & 400   & 5.60E-04                      & 283.58  \\
		& HPKM-PINN  &                & 1321    & 2.75E-04 (\textbf{$\xi$=0.7}) & 353.59  \\ 
		\hline
		\multirow{3}{*}{\begin{tabular}[c]{@{}c@{}}Convection\\-diffusion\end{tabular}} & PINN      & [2,20,20,20,1] & 921   & 9.37E-01       & 75.17   \\
		& PIKAN        & [2,5,5,1]      & 400  & 9.74E-01                 & 445.96  \\
		& HPKM-PINN &                  & 1321  & 7.35E-01 (\textbf{$\xi$=0.2}) & 467.8   \\ 
		\hline
		\multirow{3}{*}{Hemholtz}             & PINN      & [2,20,20,20,1] & 921       & 2.72E-01                         & 141.22  \\
		& PIKAN          & [2,5,5,1] & 400       & 2.61E-01                      & 375.62  \\
		& HPKM-PINN      &           & 1321      & 2.32E-01 (\textbf{$\xi$=0.9}) & 516.10  \\
		\hline
		
	\end{tabular}
	\caption{A comparison of the performance of PINN, PIKAN, and HPKM-PINN in solving four typical partial differential equations (Poisson, Advection, Convection-Diffusion, and Helmholtz).}
	\label{table1}
\end{table}

\section{Discussion}\label{sec4}
In this work, we introduced the HPKM-PINN, a novel architecture designed to address the challenges of solving complex, multiscale PDEs. By combining the complementary strengths of MLP and KAN, HPKM-PINN enhances the model’s capacity to effectively capture both high-frequency and low-frequency components, thus improving its ability to represent intricate physical phenomena. The key innovation of HPKM-PINN lies in its dynamic balancing mechanism, facilitated by the tunable weight parameter $\xi$, which adjusts the relative contribution of the MLP and KAN components. This flexibility allows the model to adapt to a wide range of problem types, optimizing performance across varying solution characteristics. Through systematic experimentation, we demonstrated that HPKM-PINN outperforms both traditional PINNs and standalone KAN models in terms of accuracy, particularly when solving typical PDEs such as Poisson, Advection, Convection-Diffusion, and Helmholtz equations. The parallel structure of the model also resulted in faster convergence, showing its robustness and versatility in handling different problem scales.

Our robustness analysis, conducted under noisy conditions, further highlighted the advantages of HPKM-PINN. Despite the presence of Gaussian noise, the model maintained superior performance compared to the MLP-only and KAN-only configurations, demonstrating its resilience to real-world data irregularities. This resilience is a key factor in the broader applicability of HPKM-PINN for real-world physical problems where noise is common. Moreover, the parallel structure of HPKM-PINN led to an increased number of parameters, which resulted in longer training times and added computational complexity. This highlights the need for optimization strategies to improve the model’s efficiency, particularly for large-scale datasets or high-dimensional problems. Future work could focus on developing adaptive network structures that adjust the complexity of MLP and KAN components based on the specific task, helping to reduce unnecessary computational overhead and scale the model more effectively. 

In summary, the HPKM-PINN architecture represents a significant step forward in Physics-Informed Neural Networks, offering a robust, adaptable, and efficient framework for solving a wide variety of PDEs. While the model excels in many scenarios, ongoing research into optimization techniques, advanced architectures, and integration with domain-specific knowledge will be essential to further enhance its performance. HPKM-PINN has the potential to become a powerful tool for tackling real-world, complex physical problems across various scientific and engineering domains.

\section*{Acknowledgements}
This work was supported by the Anhui Provincial Natural Science Foundation under Grant Nos. 2408085QF211, and 2308085QF213; and partly supported by the National Natural Science Foundation of China (NSFC) under Grant Nos. 62201005, 62304001, and 62274002; and partly supported by the Natural Science Research Project of Anhui Educational Committee under Grant No 2023AH050072.


\bibliographystyle{elsarticle-num} 
\bibliography{reference.bib}
\biboptions{sort&compress}






\end{document}